\documentclass[]{article}
\usepackage[a4paper, margin=1in]{geometry} 
\usepackage[utf8]{inputenc}
\usepackage{amsmath}
\usepackage{graphicx}
\usepackage{amssymb} 
\usepackage{url}
\usepackage[hidelinks]{hyperref} %
\usepackage[usestackEOL]{stackengine}  %
\usepackage{enumitem} %
\usepackage[backend=biber,style=authoryear,
maxcitenames=2,
maxbibnames=99,
doi=true,url=false,isbn=false,dashed=false,eprint=false, %
uniquename=false,uniquelist=false,
date=year,
backref=false,
sortcites=false
]{biblatex}
\usepackage[hang,flushmargin]{footmisc} %
\usepackage{crimson} %
\usepackage[T1]{fontenc} %
\usepackage{cochineal}   %
\usepackage[cochineal,varg]{newtxmath}  %
\usepackage{placeins} %
\usepackage{caption}  %
\title{\vspace{-1cm}
	Conceptual similarity and communicative need \\ shape colexification: an experimental study
}
\author{%
\parbox{\linewidth}{%
\noindent\centering%
\hspace{-1em} 
Andres Karjus\textsuperscript{1,2,3}, 
Richard A. Blythe\textsuperscript{3,4}, 
Simon Kirby\textsuperscript{3},
Tianyu Wang,
Kenny Smith\textsuperscript{3}\\%
}%
}%
\date{\vspace{-1em}
{\small%
\textsuperscript{1}ERA Chair for Cultural Data Analytics, Tallinn University\\%
\textsuperscript{2}School of Humanities, Tallinn University\\%
\textsuperscript{3}Centre for Language Evolution, School of Philosophy, Psychology and Language Sciences, University of Edinburgh\\%
\textsuperscript{4}School of Physics and Astronomy, University of Edinburgh\\%
}%
}
\begin{document}
	\maketitle
	
	\begin{abstract}\noindent
		Colexification refers to the phenomenon of multiple meanings sharing one word in a language. Cross-linguistic lexification patterns have been shown to be largely predictable, as similar concepts are often colexified. We test a recent claim that, beyond this general tendency, communicative needs play an important role in shaping colexification patterns. We approach this question by means of a series of human experiments, using an artificial language communication game paradigm. Our results across four experiments match the previous cross-linguistic findings: all other things being equal, speakers do prefer to colexify similar concepts. However, we also find evidence supporting the communicative need hypothesis: when faced with a frequent need to distinguish similar pairs of meanings, speakers adjust their colexification preferences to maintain communicative efficiency, and avoid colexifying those similar meanings which need to be distinguished in communication. This research provides further evidence to support the argument that languages are shaped by the needs and preferences of their speakers.
		
	    \vspace{1em}
	    \begingroup
	    \noindent
        \begin{flushleft}
        \end{flushleft}
        \endgroup
	 
	\end{abstract}

	\section{Introduction}\label{sec_intro}

    When two or more functionally distinct senses are associated with a single lexical form in a given language, expressed by a single word, then these senses can be referred to as being \emph{colexified} \parencite[][]{francois_semantic_2008}. Colexification is more general than homonymy and polysemy (which also refer to a multiplicity of senses for a word), making no assumptions about the nature of the relations between the senses.
    Recognizing that a word lexifies more than one sense requires a means to determine what the minimal units of meaning are. This would be difficult to do based on just one language, but can and has been done utilizing systematic cross-linguistic comparison. For example, English has separate monomorphemic words for \textit{leg} and \textit{foot}, while many other languages use a single word to refer to the whole thing, 
	e.g. Estonian (\textit{jalg}), Irish Gaelic (\textit{cos}) and Modern Hebrew (\textit{regel}).
	This does not mean that referring to these concepts separately in these languages is impossible, but rather it may involve more complex compounds or expressions to describe them (e.g. Estonian \textit{jalalaba} `foot', lit. `wide part of the leg, leg-blade'). It does however suggest that \textsc{leg} and \textsc{foot} could potentially be a set of comparable, minimally distinct senses, which some languages colexify, while others do not.

	Colexification effectively reduces the complexity of the lexicon, or a subdomain of the lexicon, by reducing the number of words needed to cover that space. A smaller set of words is easier to learn and remember --- the associated ``cognitive cost" is lower \parencite[cf.][]{kemp_semantic_2018}.
	But if a language, or a sub-domain of language such as the body parts lexicon, becomes too simple, then miscommunication becomes more likely. If the sub-domain is very complex and precise, then this error, or ``communicative cost" is lower --- but a language can only be so complex before it becomes unfeasible to learn. 

\begin{figure}[htb]
		\noindent
		\centering
		\includegraphics[width=0.7\columnwidth]{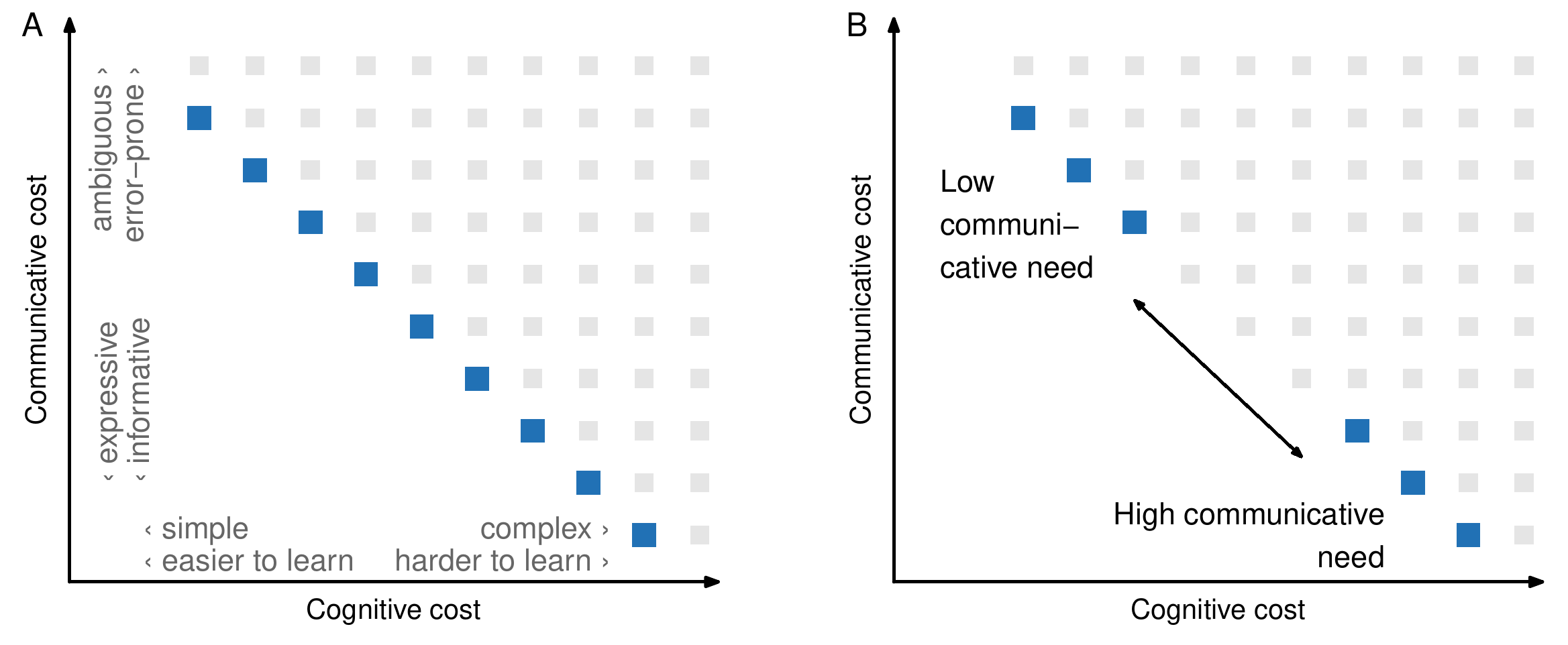}
		\caption{
			The simplicity-informativeness or cognitive-communicative cost trade-off \parencite[adapted from][]{kemp_semantic_2018}. Previous research has shown that natural languages strive to be ``as informative as possible for their level of simplicity, and as simple as possible for their level of informativeness" \parencite[][]{kemp_semantic_2018}. This interplay gives rise to the ``optimal frontier", indicated here by dark blue squares; light gray squares stand for possible but unattested languages (see panel A). 
		    It has also been argued that where exactly a language or lexical subdomain ends up along the frontier depends on communicative needs (panel B). When a language community has high communicative need to distinguish concepts in some topic or domain of the lexicon, the bias for simplicity may give way to reduce ambiguity and as such communicative cost (bottom right corner of panel B). When some domain is less relevant to a community, it may end up being simplified, at the expense of increased potential for communicative error (top left in panel B).
		}\label{fig_infcompexample}
	\end{figure}

	A successful language therefore needs to optimize these pressures to be both simple enough to be learnable \parencite[][]{gasser_origins_2004,christiansen_language_2008,smith_linguistic_2013,kirby_compression_2015}, but also meet the speakers' requirements for expressive and sufficiently precise communication. This is known as the simplicity-informativeness trade-off \parencite[see][]{kemp_kinship_2012,carr_simplicity_2020}, illustrated in Figure \ref{fig_infcompexample}A.

	\textcite{kemp_semantic_2018} argue that it is social and cultural communicative needs that modulate this perpetual optimization process. If a given domain is less relevant to users of a language, then a language may give way to the pressure of simplicity, and some sense distinctions may be lost or colexified. If something is important and part of frequent discourse, then it will likely be lexified so as to avoid error in communication --- increasing informativeness at the expense of increasing complexity (see Figure \ref{fig_infcompexample}B). 
	The topics of informativeness, complexity and communicative need have been discussed in the context of a variety of domains, including
	expressions of color \parencite[][]{lindsey_color_2002,gibson_color_2017,zaslavsky_color_2019}, 
	kinship \parencite[][]{kemp_kinship_2012}, pronouns \parencite[][]{zaslavsky_lets_2021},
	numeral systems \parencite[][]{xu_numeral_2020}, natural phenomena \parencite[][]{berlin_ethnobiological_1992,regier_languages_2016,zaslavsky_semantic_2020,kemp_season_2019}, writing systems \parencite[][]{hermalin_efficient_2019,miton_graphic_2021}, and various morphosyntactic features \parencite[][]{haspelmath_explaining_2017,mollica_grammatical_2020}. 
	Similar adaptation and efficiency effects have also been observed in experimental settings with artificial languages \parencite[cf.][]{fedzechkina_language_2012,winters_languages_2015,tinits_usage_2017,nolle_emergence_2018,chaabouni_communicating_2021,guo_expressivity_2021}.

	In a recent study based on a large sample of colexifications from a database \parencite[][]{borin_intercontinental_2013} of about 250 languages, \textcite{xu_conceptual_2020} demonstrate that similar and associated senses (like \textsc{fire} and \textsc{flame}) are more frequently colexified than unrelated or weakly associated meanings (like \textsc{fire} and \textsc{salt} for example), suggesting that this provides an important constraint on the evolution of lexicons.
	This work follows a line of research on the variability of lexification patterns across languages of the world \parencite[e.g.][]{malt_knowing_1999,francois_semantic_2008,list_using_2013,majid_semantic_2015,srinivasan_how_2015,thompson_quantifying_2018}.
	\textcite{xu_conceptual_2020} also put forward a hypothesis that, beyond the tendency to colexify similar senses, language- and culture-specific communicative needs should be expected to affect the likelihood of colexification of similar concepts --- such as \textsc{sister} and \textsc{brother}, or \textsc{ice} and \textsc{snow} --- if it is necessary for efficient communication to distinguish them.
	The latter pair in particular was investigated by \textcite[][]{regier_languages_2016}, who used multiple sources of linguistic and meteorological data to show that languages spoken in colder climates are statistically more likely to distinguish \textsc{ice} and \textsc{snow}, while languages spoken in warmer climates are more likely to colexify them \parencite[thus providing an empirical test to ideas going back to][]{sapir_language_1912,whorf_science_1956}. They argued this to be an example of lexicons being shaped by local cultural communicative needs, in this case which are in turn shaped by local physical environments.

	Languages investigated in cross-linguistic typological studies \parencite[like][]{francois_semantic_2008,regier_languages_2016,xu_conceptual_2020} have evolved to be the way they are over time, through incremental changes in how generations of speakers produce utterances by assigning signals to meanings. In this paper, we employ an artificial language experimental setup to probe lexification decisions by individual speakers, with the aim of investigating the discourse-level generative mechanisms that in natural languages would eventually lead to the observed cross-linguistic patterns.

	In Experiment 1 we investigate how colexification choices play out in a dyadic communicative task when communicative needs are uniform, and confirm that in this neutral condition, speakers do indeed prefer to colexify similar concepts. Comparison of this baseline to a condition where colexification of similar meanings would impede communication shows a reduced tendency to colexify similar meanings, providing evidence in line with the hypothesis proposed by \textcite[][]{xu_conceptual_2020}, that communicative needs of speakers modulate colexification dynamics beyond conceptual similarity. 
	
	In Experiment 2 we replicate Experiment 1 with a crowdsourced participant sample. We continue using crowdsourcing to test a variant of the original hypothesis in Experiments 3, and in Experiment 4 relax the constraints on the signal space we provide our participants in order to further explore how communicative need operates under reduced constraints on complexity. The results of these follow-up experiments support the findings of Experiment 1.
	The implications of the findings and pathways to future research are further discussed in Section \ref{section_colex_discussion}.

	\section{Experimental methodology}\label{sec_general_methodology}
	
	All our experiments use a dyadic computer-mediated communication game setup \parencite[cf.][]{scott-phillips_language_2010,galantucci_experimental_2012,winters_languages_2015,kirby_compression_2015} to investigate how similarity and communicative need interact to shape colexification choices by language users.
	In this section, we first provide a general overview of the methods and analysis  we used in all four of our experiments, before setting out each experiment in turn in more detail in later sections; unless stated otherwise, the setup for conducting a given experiment is as described in the general methods section here.
	
	In all our experiments, pairs of participants are faced with the task of communicating single-word messages using a small set of artificial words. In order to successfully do so, they must initially negotiate the meanings for the signals through trial and error. 
	The task was introduced to participants as an ``espionage game" where the usage of secret codes is justified as keeping the messages hidden from the enemy.
	The experiment interface was implemented as a web app in the R Shiny framework \parencite[][]{chang_shiny_2020}.

	\subsection{Procedure}\label{sec_expm_procedure}
	
	All our experiments consist of 135 rounds. At each round, one participant is the {\em sender} and the other is the {\em receiver}; these roles switch after every round. The sender is shown two meanings, represented by English nouns (see Section~\ref{sec_exp_stims_meanings}), and is instructed to communicate one of those meanings to the receiver, using a single word from an artificial lexicon (see Section~\ref{sec_exp_stims_signals}). The receiver is then shown the same pair of meanings (in random order) and the sender's signal, and has to guess which of the two meanings the signal represents. After taking a guess, both participants are shown an identical feedback screen which informs them whether the receiver guessed correctly. See Table~\ref{table_expm_screens} for an illustration. The game ends with a screen showing both participants their total score, asking them for optional feedback, and leaving them with instructions on how to claim their monetary reward.
	
		\begin{table}[htb]
		\centering
		\renewcommand{\arraystretch}{1.1}
		\begin{tabular}{l|l|r}
			\hline
			Player 1 & Player 2 & Comment \\ 
			\hline
			\Longunderstack[l]{ 
			 \textsc{motor} \; \textsc{essay} \\
			 Communicate \textsc{motor} using...\\
			 \textbf{nepa qohe lali fuwo nire ruqi lumu}  
			 } &  
			 \Longunderstack[l]{
			 \textsc{essay} \; \textsc{motor} \\ 
			 Waiting for message...
            } &
            \Longunderstack[r]{
            Round 1 starts \\
            Player 1 is the sender,\\ clicks on ``fuwo"} \\
            \hline
            \Longunderstack[l]{ 
            Sent \textsc{motor} using fuwo \\
            Stand by...
            } & 
            \Longunderstack[l]{
			 \textsc{essay} \; \textsc{motor} \\ 
			 Message: fuwo \\
			 This means: \textbf{ \textsc{essay} \; \textsc{motor} } \\
            } &
            \Longunderstack[r]{
            Player 2 (receiver) \\ clicks on ``\textsc{motor}"  
            } \\
            \hline
            Correct guess! \; \textsc{motor} = fuwo &
            Correct guess! \; \textsc{motor} = fuwo & Feedback screen \\
			\hline
			 \Longunderstack[l]{
			 \textsc{threat} \; \textsc{purse} \\ 
			 Waiting for message...
            } &
            \Longunderstack[l]{ 
			 \textsc{threat} \; \textsc{purse} \\
			 Communicate \textsc{threat} using...\\
			 \textbf{nepa qohe lali fuwo nire ruqi lumu} 
			 } & 
            \Longunderstack[r]{
            Round 2 starts \\
            Player 2 is the sender,\\ clicks on ``qohe"} \\
            \hline
		\end{tabular}
		\caption{
			An example of how the communication game works. Each row corresponds to a change in the screens displayed to the players, the cells in the columns indicate what is currently being shown to each of the two Players. Interactive elements of the graphical user interface are highlighted here in bold font. In Round 1, it is Player 1's turn to signal and Player 2's turn to guess the meaning of the signal. The meanings, here \textsc{motor} and \textsc{essay}, are being displayed to both Players, but in randomized order. After the feedback screen has been is shown (informing the players if the guess was correct), the roles are switched, and the next round begins.
		}\label{table_expm_screens}
	\end{table}
		
	The participants never see more than two meanings on the screen at any time. In each game, there are 10 meanings, among them 3 ``target pairs" consisting of two highly similar meanings, e.g. \textsc{motor} and \textsc{engine}. The remaining 4 meanings serve as distractors that have low similarity scores to all other meanings, including the targets (see below for details). The signal space consists of 7 artificial words such as \textit{fuwo} or \textit{qohe}. Since there are fewer signals than meanings, participants must colexify some meanings.
	We assume it takes a while to establish stable meaning correspondences, so we consider the first $1/3$ of the rounds as a ``burn-in" phase, and only analyse the final 90 rounds of the experiment.

	\subsection{Stimuli: meanings}\label{sec_exp_stims_meanings}
	
	The meanings to be communicated are English common nouns of 3 to 7 characters in length, drawn from the Simlex999 dataset \parencite[][]{hill_simlex999_2015}, which consists of pairs of words and their crowdsourced similarity judgments. We use Simlex999, as it was built for evaluating models of meaning with the explicit goal of distinguishing genuine similarity (synonymy) from associativity \parencite[the dataset incorporates a subset of the USF Free Association Norms for that purpose, cf.][]{nelson_university_2004}. Our target pairs are required to have a Simlex similarity score of at least 8 out of 10, but a free association score below 1 out of 10. This should yield meanings which are near-synonymous and not simply contextually associated. 
	For example,  \textit{arm} and \textit{leg} have a high association score of 6.7, i.e. when people hear \textit{arm} they're likely to think of \textit{leg} --- but these two are not synonyms (reflected in the Simlex similarity score of only 2.9). 
	\textit{plane} and \textit{jet} have high similarity (8.1), but they also appear to be highly associated terms (6.6).
	In contrast, \textit{abdomen} and \textit{belly} have high similarity (8.1) but at the same time low associativity (0.1), therefore making them suitable for our stimulus set.
		 
\begin{figure}[htb]
		\noindent
		\centering
		\includegraphics[width=1\columnwidth]{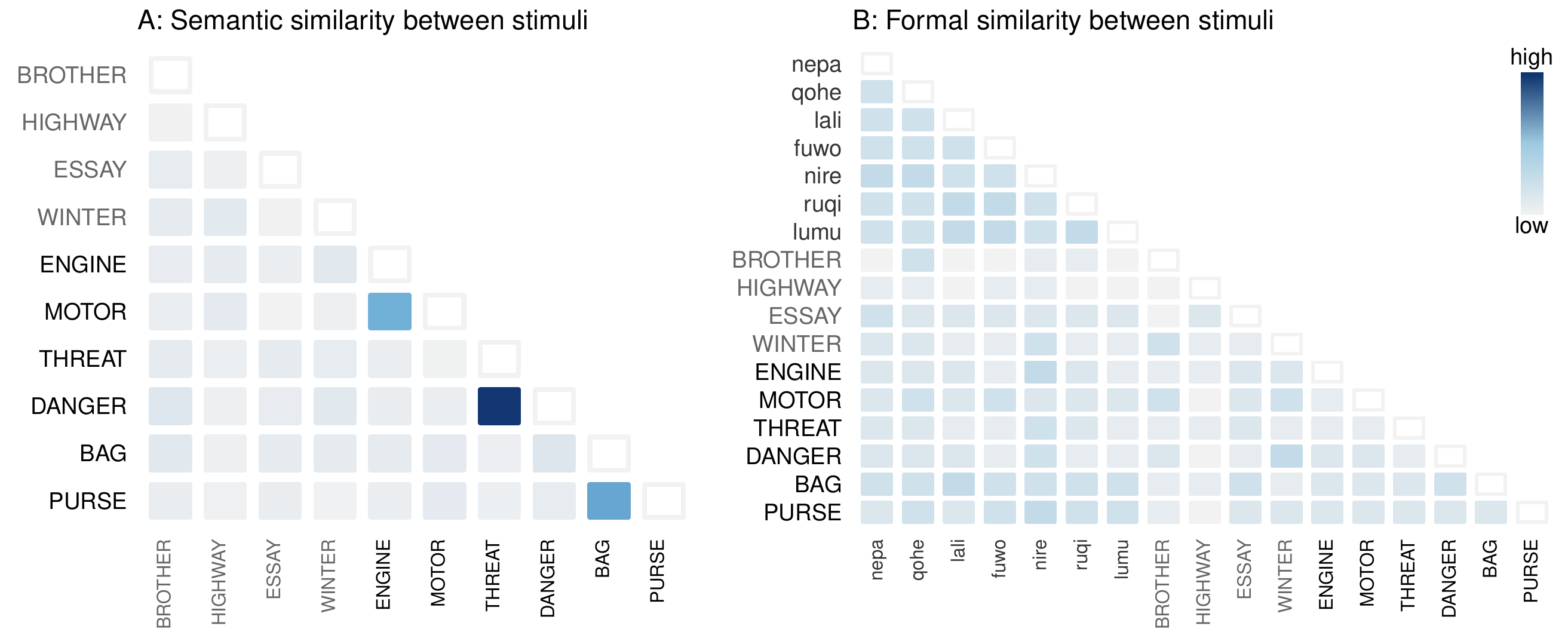}
		\caption{
			The meanings and signals used in one game (dyad no. 21 in Experiment 1). The left side A panel illustrates the meaning space: only the target pairs have high similarities (dark blue), with low similarities between all other meanings. The diagonal (self-similarity) is marked by white squares.
			The right side B panel shows the similarity of form (as inverse of edit distance), of both the meanings (in caps) and signals (lowercase). 
			The stimuli in our experiments are generated in a way that ensures only target meanings are semantically similar to one another, and that form similarity remains low across the board.
		}\label{fig_expmap}
	\end{figure}

	Since Simlex does not have scores for all possible word pairs in its lexicon, we also used publicly available pretrained word embeddings \parencite[fasttext trained on Wikipedia, cf.][]{bojanowski_enriching_2017} to obtain additional computational measures of similarity. We use these to ensure low similarity across the board in the distractor set, which was sampled so that no two distractors and no distractor-target pair would have vector cosine similarity above 0.2 (out of 1). Furthermore, no two nouns were substrings of each other, nor otherwise similar in form (we used a Damerau-Levenshtein edit distance\footnote{
	    A metric that corresponds to the the minimum number of operations required to change one word into the other. The operations consist of insertions, deletions, substitutions of single characters, and as an addition to the original Levenshtein distance, the transposition of two adjacent characters. The edit distance to get from \textit{bag} to \textit{purse} would be 5: 3 substitutions plus the \textit{s} and \textit{e} inserted at the end. 
	} threshold of 3), and targets were not allowed to share the same first letter. 
	This filtering yielded 13 eligible target pairs: \textsc{abdomen-belly, area-zone, author-creator, bag-purse, coast-shore, couple-pair, danger-threat, drizzle-rain, engine-motor, fashion-style, job-task, journey-trip, noise-racket}.
	The meaning space for each dyad consists of 3 target pairs selected at random from this set of 13 possible pairs, plus an additional 4 distractor meanings, drawn from the remaining 475 possible meanings (see Figure~\ref{fig_expmap}.A).

	\subsection{Stimuli: signals}\label{sec_exp_stims_signals}
	
	For each dyad in Experiments 1-3, we generated a set of 7 signals, according to the following constraints (Experiment 4 has an extended signal space of 10 signals). Each signal had a length of 4 characters, and was composed of 2 consonant-vowel syllables, constructed from a set of consonants $\{ qwtpsfhnmrl \}$ and vowels $\{ aeoui \}$. We further constrained the artificial language so that the initial letters of the signals would not overlap with any initial letters of the meanings (English nouns) in a given stimulus set. We used a large English word list to make sure there was no overlap between the artificial signals and actual English words, and furthermore made sure all signals were at least 3 edits distant from the meanings (the English nouns) in the same game, as well as from other signals in the game (see Figure~\ref{fig_expmap}.B).
	
	\FloatBarrier

	\subsection{Experimental manipulation of communicative need}\label{sec_exp_conditions}
	
	The experiments have two conditions: the {\em baseline} or control condition, where communicative need is uniform, and the {\em target} condition where we manipulate communicative need, creating a situation where colexifying certain (similar) concepts would hinder the accurate exchange of messages. This applies to all experiments except for Experiment 3 which only has a (modified) target condition (details in Section \ref{sec_expm3}).
	
	In the baseline condition, the distribution of meaning pairs (e.g. \textsc{drizzle-rain}, \textsc{style-fashion}, \textsc{payment-bull}, \textsc{rain-payment}, \textsc{rain-fashion}) is uniform --- each possible combination is shown to the participants exactly 3 times.
	Based on cross-linguistic tendencies \parencite[cf.][]{xu_conceptual_2020}, in this condition we would expect participants to colexify similar meanings.
	In the target condition, all possible meaning pair combinations still occur in the game, but crucially, we manipulate the occurrence frequencies so that the target (similar-meaning) pairs occur together more often than the distractor pairs.
	The target pairs (e.g. \textsc{drizzle-rain}, \textsc{style-fashion}) are shown 11 times each, and the pairs consisting of distractor meanings (e.g., \textsc{payment-bull}) 5 times each. Pairs consisting of a meaning from a target pair plus another meaning are shown 2 times (e.g. \textsc{rain-payment} or \textsc{rain-fashion}).
	
	The non-uniform distribution of meaning pairs in the target condition entails that to communicate successfully, participants are required to select signals which allow their partner to differentiate between \textsc{drizzle} and \textsc{rain} 11 times, but are only required to differentiate between \textsc{rain} and \textsc{payment} 2 times. The increased co-occurrence of similar meanings in the target condition simulates communicative need. If a pair of similar meanings never or seldom needs to be distinguished, then it is efficient to colexify them, both from a learning and communication perspective. In contrast, if the communicative context often requires disambiguating between two similar meanings or referents --- such as \textsc{rain} and \textsc{drizzle} in a culture obsessed with talking about poor weather --- then colexifying them as \textit{rain} or blending them into something like \textit{rainzzle} would obviously be detrimental to communicative success. We expect this to be reflected in the outcomes of the target condition: participants should avoid colexifying the similar concepts, as that would make it difficult to distinguish them and hinder communicative efficiency. 
	
	These pairs are displayed in a randomized order, but randomized separately for the burn-in (first third) and post-burn-in part of the game. In other words, we make sure that if \textsc{rain-fashion} is supposed to appear 3 times over the course of the game, then it will appear once in the burn-in and twice afterwards. Of course not all values are divisible by 3, so the distribution of stimuli between burn-in and the rest of the game is not perfect, but we optimize it to be as good as numerically possible.\footnote{The randomized order could in principle cause a situation where some meaning comes up in multiple successive rounds, or, on the contrary, does not appear for a while. However, since there are only 10 meanings in each game, such runs should be relatively rare and therefore have limited systematic affect on our results.}
	
	The meaning pair frequency distribution used in both conditions ensures that individual meanings all occur exactly the same number of times (which is 27). 
	It is necessary to control for individual frequencies in this manner, as simply making target pairs more frequent would also mean making the individual meanings in those pairs more frequent than the distractors. This would introduce a confound: another reasonable hypothesis could be that it is occurrence frequency that drives colexification \parencite[i.e. colexifying frequent meanings is preferred, or avoided; cf.][]{xu_conceptual_2020}, over and above communicative need or word similarity.

\subsection{Participants and exclusion criteria based on communicative accuracy}\label{sec_exp_participants}
    
    All our experiments were approved by the Ethics Committee of the School of Philosophy, Psychology and Language Sciences of the University of Edinburgh.
    All participants provided informed consent on the online game platform prior to participating and were compensated monetarily for their time.
    We do not include all the data in our analysis, as the communicative accuracy of some dyads is at or near random chance. Low accuracy in guessing the meanings of the signals transmitted indicates a given dyad did not manage to converge on a lexicon they could use with reasonable success to communicate --- such data are not informative in terms of our research question.
    
	We calculate the communicative accuracy of a dyad as the percentage of correct guesses out of all guesses made in the game after the burn-in period (see Section~\ref{sec_expm_procedure}). We make the assumption that dyads with a total accuracy score above 59\%, i.e. guessing correctly in $53/90$ of the trials, were unlikely (binomial $p = 0.036 < 0.05$) to be signaling or guessing randomly. Dyads scoring below this threshold were excluded from analysis; all excluded dyads still received full payment. The instructions also included a request not to make any notes or write anything down during the experiment, and participants were asked at the end of the experiment whether they had taken written notes; no participants were excluded on the basis of having admitted to taking written notes.

\subsection{Quantifying colexification}\label{sec_exp_operationalize}
    
\begin{figure}[htb]
		\noindent\centering
		\includegraphics[width=0.75\columnwidth]{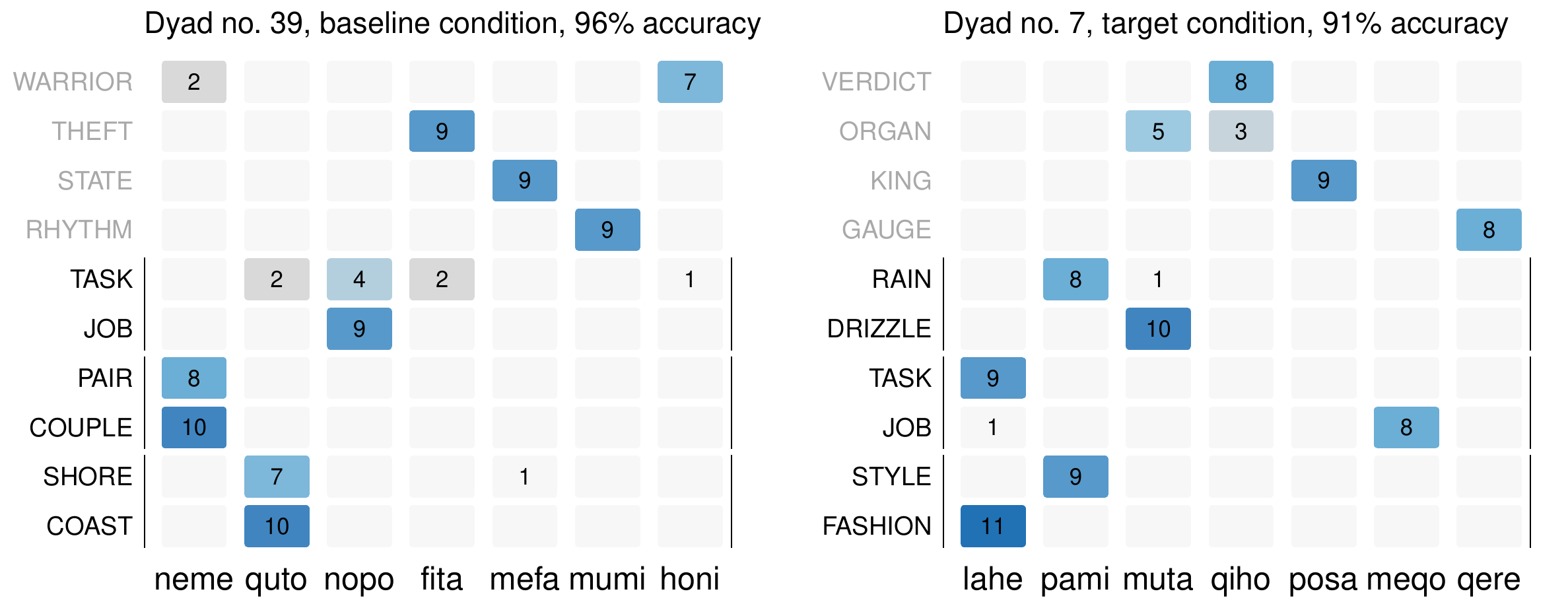}
		\caption{
			Example signal-meaning matrices from two games in Experiment 1.
			The vertical black bars highlight the similar-meaning target pairs.
			Counts in the cells indicate how many times in total each signal was used to communicate each meaning in a given game (larger values have darker cells).
		    In the baseline condition game (left), the players have chosen to colexify similar meanings such as \textsc{shore} and \textsc{coast}.
			In the target condition one (right), similar meanings often need to be distinguished from one another. The players have responded to this pressure by colexifying \textsc{rain} with \textsc{style} and \textsc{drizzle} with \textsc{organ} instead. 
		}\label{fig_expm_matrices}
	\end{figure}

    We are interested in what participants do with the target meaning pairs. If they colexify target meanings in the baseline condition (e.g. use the same signal for \textsc{rain} and \textsc{drizzle}), that would support the cross-linguistic findings of \textcite[][]{xu_conceptual_2020}, that similar meanings are most often colexified. If participants avoid colexifying target meaning pairs in the target condition, then this would support the hypothesis we intend to test, that given high enough communicative need to distinguish similar meanings, they will not be colexified (see Figure \ref{fig_expm_matrices} for an illustration).
    We therefore need a method to operationalize these signal-meaning associations in a manner that would allow us to test our hypothesis via rigorous statistical analysis, while preferably also capturing possible changes in lexification choices over the course of each game.
    
	Data from each game is converted into a new dataset suitable for statistical analysis, consisting only of colexifications involving one of the target meanings. We introduce a new binomial variable, ``colexification with synonym", and assign values to it using the following procedure. 
	We start by iterating through all the messages sent in the post-burn-in part of the game that signal one of the target meanings. We check if the most recent usage of a given signal by the same player in the same game lexified another meaning, and if that meaning belongs to the same target pair or not.
	For example, given Player 1 signals \textsc{rain} using \textit{pami} at round number 53, we check the most recent usage of \textit{pami} by Player 1 in all preceding rounds of the game where Player 1 was the signaler. Let's say this happened at round 39, and Player 1 used \textit{pami} to also signal \textsc{rain}. This does not count as a colexification --- it's the same meaning --- so this case is not added to the new dataset.
	At some point later in the game, at round 127, Player 1 signals \textsc{rain} using \textit{pami} again. The previous usage of \textit{pami} by Player 1 was round 121, where they used it to signal \textsc{style}. This counts as a colexification, and so this case is added to the new dataset. 
	As \textsc{style} is not a synonym of \textsc{rain}, the value assigned to the new ``colexification with synonym" variable is ``no". If instead of \textsc{style}, the most recent meaning signaled by \textit{pami} had been \textsc{drizzle} --- a highly similar meaning belonging to the same target pair with \textsc{rain} --- then this value would be set as ``yes".
	
	This procedure is repeated for all dyads that score above our previously described accuracy threshold.
	We do not include the data from the burn-in period (first $1/3$ of the game) in the statistical analysis of the results, but do take the burn-in into account when checking for most recent usage of signals, as players do not start from a clean slate after the end of the burn-in, but already have some experience with the language and likely at least some signal-meaning associations in place. 	
	This is not the only way to operationalize and analyze such data, but it does provide insights into participant behavior while allowing for explicit comparison of the conditions in terms of the likelihood of target pair colexification, and for modeling changes in lexification over the course of the game. 
	
	The distributions of meanings occurring in the game --- the input data for the participants --- are carefully balanced, and all games consist of exactly 135 rounds, as described in Section \ref{sec_exp_conditions}. Note however, that under this operationalization, the amount of output data yielded by different dyads varies somewhat (see Figure \ref{fig_expm_tiles}). The exact number of data points per dyad depends on their colexification behavior: if they converge on a system where most of the target meanings get their own unique signal, then there will be fewer colexifications and as such fewer data points. If they colexify the target meanings, either with their near-synonyms or with unrelated meanings, then there will be more colexifications to analyze. This is by design, as we prioritize balancing the input, and is controlled for in our statistical analyses below, ensuring that the overall results are not driven by a few data-rich dyads.

	\subsection{Statistical modeling approach}\label{sec_exp_regression}
	
	We used mixed effects logistic regression models \parencite[using the lme4 package in R;][]{bates_fitting_2015} to model all the datasets derived using the approach described in Section \ref{sec_exp_operationalize} above. The model has the same effects structure across all experiments.
	Condition (baseline or target) is treatment-coded, with the baseline condition set as the reference level (with the exception of Experiment 3; this is discussed in the relevant section). Round number (scaled to a range of $[-1,1]$) is centered at round 68, the middle of the game.
	The binomial colexification variable is set as the response, predicted by condition, round number, and their interaction. The interaction with round accounts for possible changes in lexification preferences. To account for the repeated measures nature of the data and the fact that the number of data points per dyad is variable, we set random intercepts for meaning and sender (the latter nested in dyad) and a random slope for condition by meaning.\footnote{
	    In lme4 syntax: 
	    \texttt{ colexification  {\raise.17ex\hbox{$\scriptstyle\mathtt{\sim}$}} condition * round + (1+condition|meaning) + (1|dyad/sender)
	    }
	    
	} 
	A full random effects structure would be desirable, but could not be included due to model convergence issues.

\section{Experiment 1}\label{sec_expm1}

    \subsection{Methods}
    
\begin{figure}[htb]
			\noindent
			\includegraphics[width=1\columnwidth]{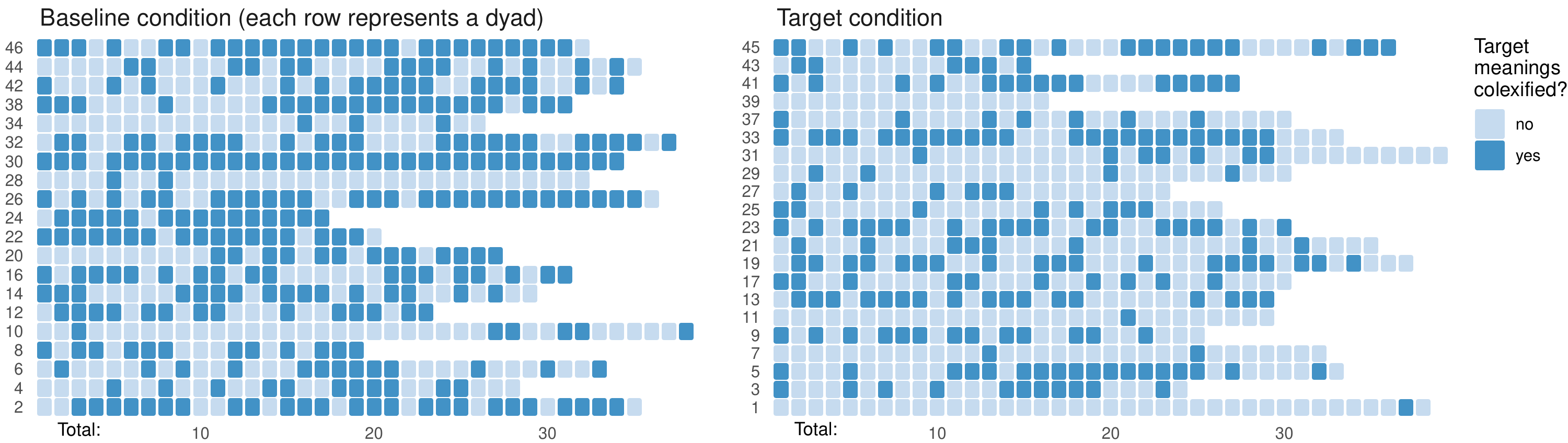}
			\caption{
				The colexification data across all dyads in Experiment 1, operationalized using the procedure described in Section \ref{sec_exp_operationalize}
				Each row is a dyad, labeled by its number. 
				Each tile corresponds to a message that contains a target meaning, which is being colexified with another meaning using the same signal: dark blue if with a related meaning, e.g. \textsc{rain-drizzle}, light blue if with an unrelated meaning. 
				Trials are shown in order, but those involving distractor meanings, or where the given signal was last used to lexify the same meaning, are excluded (therefore the final number of data points per dyads vary, despite all games having the same 135 rounds).
				The difference between conditions is visually apparent: the baseline condition on the left has more dark blue, indicating colexification of similar meanings, while the target condition on the right has more light blue tiles, indicating colexification of dissimilar meanings. 
			}\label{fig_expm_tiles}
		\end{figure}

    The setup of our first experiment matches the overview in Section \ref{sec_general_methodology}: there is a baseline and a target condition, and in both cases participants are provided with 7 signals which they can use to lexify the 10 meanings. We operationalize the data for statistical modeling using the procedure as described in Section \ref{sec_exp_operationalize} and illustrated in Figure \ref{fig_expm_tiles}.
    
	\subsubsection{Participants}
	
	The pool of participants for Experiment 1 consists of students of the University of Edinburgh, recruited though the university's CareerHub portal and departmental mailing lists.\footnote{
	    The experiment was initially planned as an in-person lab study that would have commenced in March 2020. The Covid-19 pandemic rendered this impossible, so we reworked it into an online experiment, but continued with the originally intended participant pool of student recruits. 
	    We made use of SimpleSignUp \mbox{(https://github.com/jwcarr/SimpleSignUp)} to schedule the participants.
	    For the later experiments we moved to using participants sourced from Amazon Mechanical Turk.
	}
	Participants were only allowed to complete the game once. All participants identified as native or near-native speakers of English.
	$46$ dyads finished the experiment, $41$ of which were included in the analysis (20 baseline, 21 target condition dyads; 82 participants in total). 
	Data from 5 dyads were discarded, either because they explicitly admitted to misunderstanding the game instructions in the feedback form (1 dyad) or because of low communicative accuracy that likely resulted from random guessing (4 dyads; see Section \ref{sec_exp_participants} above for discussion of our exclusion criterion). 
	
	\subsection{Results}
	
\begin{figure}[htb]
			\noindent
			\includegraphics[width=1\columnwidth]{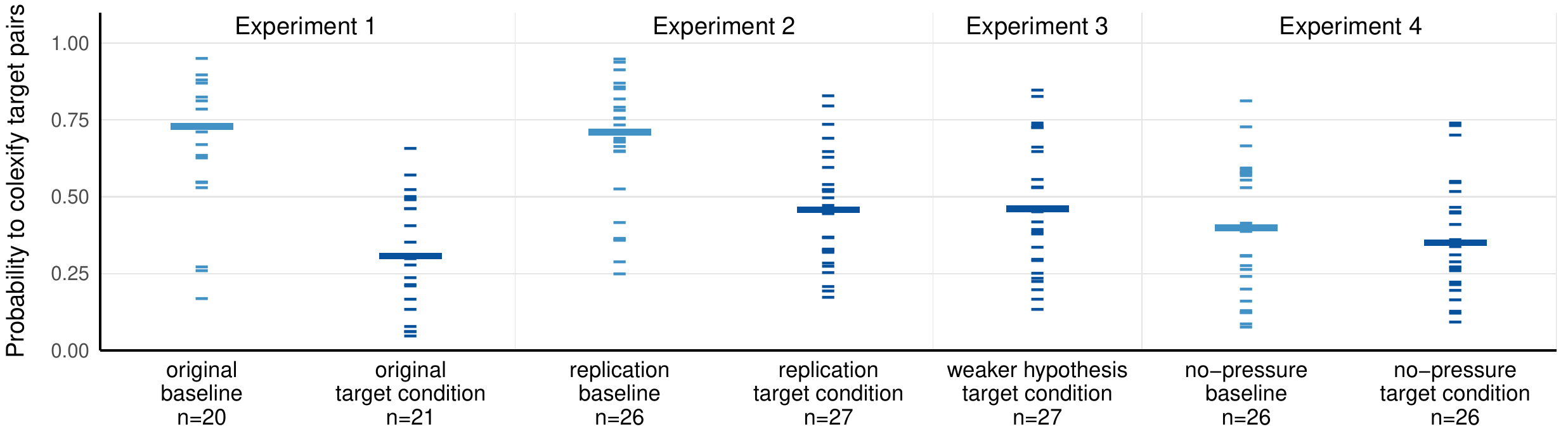}
			\caption{
				Estimated probabilities of colexifying similar-meaning target pairs by the end of the game.
				Each notch is one dyad, the wider bars are medians (unlike in our full statistical analysis, meanings are collapsed here within dyads for clearer visualization purposes). 
				Target conditions (with manipulated communicative need) are in darker blue. In Experiment 1 (left) as well as its replication (Experiment 2), dyads were more likely to end up colexifying similar meanings like \textsc{trip} and \textsc{journey} in the baseline condition, but less so in the target condition, when faced with communicative need to distinguish them. 
				This graph displays all dyads across all experiments (173 dyads or 346 participants in total) and will be referred back to in later sections.
			}\label{fig_expm_all}
	\end{figure}

    The data processing procedure described in Section \ref{sec_exp_operationalize} yields a dataset of $1214$ cases ($597$ in the baseline, $617$ in the target condition), a median of $30$ per dyad (illustrated in Figure~\ref{fig_expm_tiles}).
    In the mixed effects regression model (see Section \ref{sec_exp_regression}), the dependent variable is the derived binomial colexification measure. The fixed effects consist of the condition, round number, and the interaction of the two. The interaction is included to account for possible changes over the course of the game (cf. Figure~\ref{fig_expm_tiles}). Random effects are included to control for repeated measures of meanings, dyads and players, as outlined in Section \ref{sec_exp_regression}.
	In the model described in Table \ref{table_regression}, the intercept value of $-0.22$ stands for the log odds of target meaning pairs being colexified, in the baseline condition, mid-game (i.e. a probability of $0.45$; recall that we centered round number). By mid-game, the model is not picking up a significant difference between the conditions ($p=0.3$). Each passing round does increase the probability of colexification in the baseline condition ($\beta=1.02$, $p=0.0001$). Importantly, the interaction between condition and round is in the opposite direction ($\beta=-1.18$, $p=0.0015$), indicating participants were less likely to colexify related meanings in the target condition, where these related meanings frequently co-occurred and needed to be distinguished from one another. By the end of a game, the estimated average probability of colexifying target pairs is only $0.29$ in the target condition, compared to $0.69$ in the baseline condition. In short, these results support our communicative need hypothesis (for further illustration, see Figure \ref{fig_expm_all}, the leftmost columns).
		
	\begin{table}[htb]
		\centering
		\begin{tabular}{rrrrr}
			\hline
		colexification \textasciitilde & Estimate & SE & $z$ & $p$ \\ 
			\hline
			intercept (baseline condition, mid-game) & 	-0.22 & 0.37 & -0.59 & 0.56 \\
			+ condition (target) &                     	-0.52 & 0.51 & -1.03 & 0.3 \\
			+ round &                                  	1.02 & 0.27 & 3.84 & $<$0.01 \\
			+ condition (target) $\times$ round &      	-1.18 & 0.37 & -3.17 & $<$0.01 \\ 
			\hline                

		\end{tabular}
		\caption{
			The fixed effects from the mixed effects regression model applied to data from Experiment 1, predicting the value of the colexification variable (reference level: ``no") by the interaction between condition (reference: baseline) and round number. The latter is included to account for progress over the course of the experiment.
			The results indicate a statistically significant difference in participant behavior between the two conditions, supporting the hypothesis that communicative need can drive lexification preferences above and beyond conceptual similarity.
		}\label{table_regression}
	\end{table}

	\subsection{Discussion}\label{sec_expm1_discuss}
	
	Human memory is not infinite, and neither is time that can be used for learning. An unlimited number of signals or signal-meaning associations cannot be stored in the brain. We emulated these natural conditions in our artificial communication experiment by providing the participants with a signal space that is smaller than the provided meaning space. 
	The results from the baseline condition provide experimental support for the cross-linguistic findings of \textcite{xu_conceptual_2020} --- that people indeed tend to colexify similar meanings. Yet when faced with a situation where there is elevated communicative need to distinguish certain meaning pairs more often than others, people are more likely to colexify other pairs or clusters of meanings to maintain communicative efficiency --- even if this requires colexifying unrelated meanings. This in turn supports the hypothesis suggested by \textcite{xu_conceptual_2020}, and is in line with previous research on communicative need in general (cf. Section~\ref{sec_intro}). 
	
\begin{figure}[htb]
			\noindent
			\includegraphics[width=1\columnwidth]{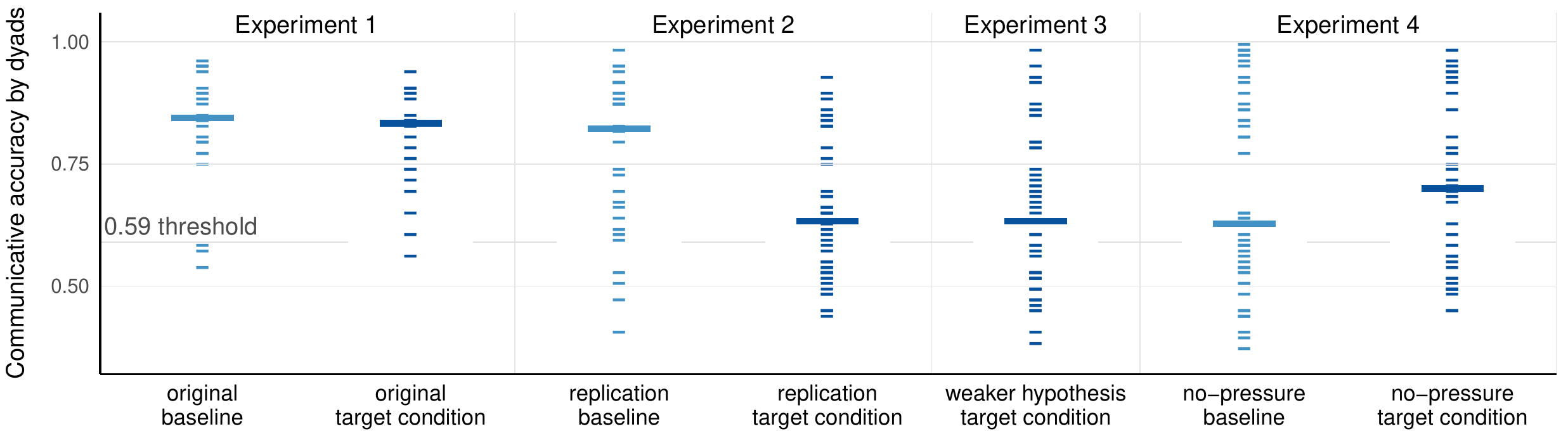}
			\caption{
				Communicative accuracy of all dyads across all experiments.
				Each notch is one dyad, the wider bars are medians. 
				Target conditions (those with manipulated communicative need) are in darker blue.
				The threshold of 0.59, that we set to filter out dyads which likely played by the random button smash strategy, is shown as a gray segmented line. The student dyads (Experiment 1) had on average higher accuracy than the crowdsourced participants in the rest of the experiments, but all conditions included some dyads that scored very low, as well as some that scored very high.
			}\label{fig_accs}
		\end{figure}

	It should be noted that this setup possibly puts a heavier cognitive load on the participants in the target condition. In the baseline condition, participants can colexify similar meanings (like \textsc{rain} and \textsc{drizzle}) without paying an additional communicative cost. It is probably safe to assume similarity-driven pairings are easier to remember. In the target condition, participants are encouraged to colexify meanings which are maximally dissimilar by design (e.g., \textsc{dentist} and \textsc{fashion}).\footnote{
		However, note that since the guess is always made just between two options, the random baseline is still 50\%. This means it is still possible to achieve a reasonably high communicative accuracy score even if the dyad decides to colexify all target meaning pairs --- as some dyads indeed do --- and just guess when a target pair comes up. Assuming otherwise perfect 100\% performance on all other, non-target pairs (this is however never observed) and that 50\% of the random guesses hit the mark, it would be technically possible to achieve 82\% accuracy in the post-burn-in part of the game, using this strategy.
	}
	In that sense, Experiment 1 constitutes a strong test of the communicative need hypothesis --- we predict that given high enough communicative need, speakers would even colexify unrelated meanings rather than sacrifice communicative efficiency.
	However, actual differences in average communicative accuracy in Experiment 1 turn out to be negligible.\footnote{
		The estimated probability of making a correct guess is 0.84 in the target condition compared to 0.88 in the baseline ($p=0.12$), based on a mixed effects logistic model, predicting correctness of guess by condition, with random slopes and intercepts for meaning and dyad (only taking into account the post-burn-in part of each game, and excluding dyads with overall accuracy below 59\%, as described above).
		}
	 Figure \ref{fig_accs} illustrates this, as well as the accuracy levels of the rest of the experiments, which we will discuss further in the next sections.
	 There was also no difference in game length (on average 29 minutes in both conditions).

	Regardless, we will still explore the weaker hypothesis in a follow-up experiment, where participants are provided with less frequently co-occurring but still similar meanings to colexify (Experiment 3 in Section \ref{sec_expm3}). 
	We will also describe an experiment with an expanded signal space (Experiment 4 in Section \ref{sec_expm4}). However, in the next section, we will first replicate the initial study on a different, slightly larger sample of participants.
		
		\FloatBarrier

		\section{Experiment 2}\label{sec_expm2}
		
		Experiment 2 is a replication of Experiment 1, where we recruited participants from Amazon Mechanical Turk; all other details of the experimental design and analyses are the same. As our replication below demonstrates, the behavior of these two samples in terms of the research question is very similar, but the samples differ somewhat in terms of average communicative accuracy.

		\subsection{Methods}
		
		\subsubsection{Participants}\label{sec_expm2_participants}
		
		We restricted participation to Mechanical Turk workers based in the United States to have a sample roughly comparable sample to Experiment 1 (i.e. largely English-speaking).
		In this and the following experiments, we only accepted workers with a history of at least $1000$ successfully completed tasks and a $97\%$ or higher approval rate. Furthermore, we used the qualifications system of Mechanical Turk to make sure no worker participated more than once (within a single experiment or across Experiments 2--4).
		As before, all participants provided informed consent on the online game platform prior to participating and were compensated monetarily for their time.
		
		$79$ dyads finished the experiment, $53$ of which were included in the analysis (26 baseline, 27 target condition dyads; 106 participants in total). 
		Data from 26 dyads was discarded, 24 because of low communicative accuracy and 2 due to suspected cheating (see below).
        Communicative accuracy turned out to be lower on Mechanical Turk than in our student sample, with many players operating at or even below random chance (see Figure \ref{fig_accs} above in Section \ref{sec_expm1_discuss}). It is unclear to us why the rejection rate was so high, and this may reflect the difficulty of our communication task relative to other tasks our participants were completing around the same time, although anecdotally we know of other researchers who had problems with data quality on Mechanical Turk in the summer of 2020.  
        There is also a greater difference in communicative accuracy between the baseline and target condition in Experiment 2, compared to Experiment 1, possibly due to the heavier cognitive load of the target condition compared to the baseline, discussed in Section \ref{sec_expm1_discuss}.
        
        We manually flagged 2 dyads as suspicious, on the basis that they achieved near-perfect accuracy while sending apparently random signals. This could potentially be one person using two Mechanical Turk accounts to play against themselves, or two workers communicating outwith the game interface. After discovering these anomalies, we manually inspected data from high-accuracy dyads in all experiments.
		It was also not uncommon for participants recruited via Mechanical Turk to simply drop out mid-game, effectively canceling the experiment for their dyad partner as well; these dyads were treated as having withdrawn consent and their data was not analyzed.

		\subsection{Results}
		
		The analysis procedure for Experiment 2 is identical to that of Experiment 1: we operationalize colexification (see Section \ref{sec_exp_operationalize}; yielding a dataset of $1659$ cases, a median of 32 per dyad) and apply the same mixed effects logistic model (Section \ref{sec_exp_regression}). Experiment 2 successfully replicated the results of Experiment 1 with the condition-round interaction coefficient being significantly negative ($\beta=-0.66$, $p=0.03$; see Table \ref{table_regression2}; refer back to Figure \ref{fig_expm_all} above for visual comparison). This value indicates participants were again less likely to colexify related meanings in the target condition which simulated communicative need (average probability of $0.46$ by the end of the game), compared to the baseline condition, where there was no such pressure, and where participants more often colexify related meaning pairs ($0.72$ probability).
		 
		\begin{table}[htb]
		\centering
		\begin{tabular}{rrrrr}
			\hline
		colexification \textasciitilde & Estimate & SE & $z$ & $p$ \\ 
			\hline
			intercept (baseline condition, mid-game) & -0.2 & 0.29 & -0.68 & 0.49 \\
			+ condition (target) &                     -0.44 & 0.36 & -1.24 & 0.22 \\
			+ round &                                   1.14 & 0.24 & 4.82 & $<$0.01 \\
			+ condition (target) $\times$ round &       -0.66 & 0.31 & -2.11 & 0.03 \\
			\hline
		\end{tabular}
		\caption{
			The fixed effects from the mixed effects regression model applied to data from Experiment 2, predicting the value of the colexification variable (reference level: ``no") by the interaction between condition (reference: baseline) and round number. The results replicate those of Experiment 1.
		}\label{table_regression2}
	\end{table}

		\subsection{Discussion}
		
		The successful replication gives us additional confidence in the overarching hypothesis concerning the role of communicative need in lexification choices. We now turn to two more experiments to gain a better understanding of how communicative need and simplicity preferences shape the behavior of our participants, testing a weaker version of the communicative need hypothesis in Experiment 3 and relaxing the signal space constraint in Experiment 4.

		\section{Experiment 3}\label{sec_expm3}
		
		As discussed in Section~\ref{sec_expm1_discuss}, the target condition in Experiments 1--2 pushes participants to colexify meanings which are highly dissimilar (recall Figure \ref{fig_expm_tiles}).
		In that sense, it tested a strong version of our hypothesis, that the need to maintain communicative efficiency would outweigh the awkwardness of colexifying unrelated concepts, e.g. \textsc{bull} and \textsc{fashion}
		\parencite[which would amount to homonymy in natural languages, something which has been shown to hinder lexical retrieval and processing; cf.][]{klepousniotou_processing_2002,beretta_effects_2005}.
		Here we explore an alternative, possibly more natural target condition, where participants can avoid colexifying target meanings by colexifying non-target meanings which are similar to one another.
		
		\subsection{Methods}
		
		\subsubsection{Participants}
		
		The sample is similar to Experiment 2: we source participants from Mechanical Turk, applying the same restrictions as in Experiment 2.
		$52$ dyads finished the experiment, $27$ of which were included in the analysis ($54$  participants in total). 
		Data from $25$ dyads was discarded, 23 because of low communicative accuracy and 2 due to suspected cheating (see Section \ref{sec_expm2_participants}). The average accuracy is similar to what we observed in Experiment 2 (see Figure \ref{fig_accs}). 
		
		\subsubsection{Procedure}
		In this experiment, there is only a single, target-type condition, i.e. one where communicative need is manipulated. The meaning space still consists of 10 meanings, with 3 high-similarity target meanings pairs and 4 distractors. Crucially, here the distractors also form 2 similarity pairs. 
		For example, in Experiment 3, a meaning space might consist of \textsc{\underline{bag, purse}, \underline{drizzle, rain}, \underline{author, creator}, danger, threat, journey, trip} (target meaning pairs underlined). 
		Recall that distractor pairs only co-occur (i.e. need to be distinguished) 5 times over the course of the game, while target pairs co-occur 11 times in the target condition.
		Colexifying the distractors is now incentivized not only by their lower co-occurrence but also their semantic similarity. Doing so would allow participants to reserve more unique signals to distinguish target meanings, the ones that do co-occur often and need to be distinguished.
		All other parameters are identical to Experiments 1 and 2.

	\subsection{Results}
		
	To obtain a point of comparison, we combine the data collected in this experiment with the baseline and target condition data from Experiment 2 (totaling $2527$ cases, of those $868$ from the weaker hypothesis target condition of Experiment 3). We fit a mixed effects logistic regression model with the same structure as before, but set the new, ``weaker" target condition as the reference level for the condition effect. We find that as with the previous target condition, there is still a significant difference from the baseline condition (the condition $\times$ round interaction $p=0.002$). However, participant behavior does not differ from the target condition of Experiment 2, i.e., our manipulation of the meaning space did not elicit a meaningful difference ($p=0.29$, see Table \ref{table_regression3}, and Figure \ref{fig_expm_all} above).
		
	\begin{table}[htb]
		\centering
		\begin{tabular}{rrrrr}
			\hline
		colexification \textasciitilde & Estimate & SE & $z$ & $p$ \\ 
			\hline
			intercept (weaker target condition, mid-game) &  -0.22 & 0.25 & -0.9 & 0.37 \\
			+ condition (baseline)  &                           0.04 & 0.34 & 0.12 & 0.9 \\
			+ condition (target) &                  -0.39 & 0.34 & -1.16 & 0.24 \\  
			+ round &                                           0.14 & 0.2 & 0.68 & 0.49 \\
			+ condition (baseline) $\times$ round             &0.95 & 0.3 & 3.11 & $<$0.01 \\
			+ condition (target) $\times$ round &      0.3 & 0.29 & 1.06 & 0.29 \\   
			\hline
		\end{tabular}
		\caption{
			The fixed effects from the mixed effects regression model applied to combined data from Experiments 2 and 3, predicting the value of the colexification variable (reference level: ``no") by the interaction between condition (reference: new weaker target condition) and round number. Participant behavior does not differ significantly between the original target condition and the new weaker-hypothesis target condition.
		}\label{table_regression3}
	\end{table}

	\subsection{Discussion}
		
	We once again observe the same general trend \parencite[echoing the cross-linguistic findings of][]{xu_conceptual_2020} that participants colexify any similar-meaning pairs even if it causes the occasional miscommunication, but change their behavior if that cost becomes too high, as evidenced once more by the significant difference between the baseline condition of Experiment 2 and the new target condition of Experiment 3. We also expected that participants would colexify target pairs even less often than in the target conditions of Experiment 2, since alternative similar pairs were available among the distractors, removing one of the assumed barriers to avoiding colexifying target pairs. The data does not support this hypothesis; we see the same (relatively low) rate of colexification of target pairs in our new data. One explanation may be that the difference between co-occurrence frequencies is simply not stark enough for participants to see a benefit (consciously or subconsciously) in colexifying the distractors --- and there is still the option to colexify target-distractor pairs, which co-occur less often (twice each, versus the five of the distractor-only pairs).

\section{Experiment 4}\label{sec_expm4}
        
    This follow-up to the initial study explores the effect of removing constraints on the signal space. In all the previous experiments, participants were provided 10 meanings but only 7 signals to work with. This meant some meanings would be colexified, either accidentally (leading to lowered communicative accuracy) or systematically (allowing for higher accuracy). 
    Here, we remove this requirement to colexify, with the aim of gaining a better understanding of participants behavior when this central component of our experimental setup is relaxed. 
    We expect one of two outcomes. Participants may well behave more or less the same as in Experiments 1--2, making use of a limited number of signals and colexifying some meanings --- after all, 7 signals should be easier to learn and remember than 10. Alternatively, if 10 signals is not too unmanageable, participants may forgo colexification altogether: this would eliminate the difference between baseline and target condition outcomes.

    \subsection{Methods}
    \subsubsection{Participants}
    We continue using participants from Mechanical Turk as before.
	$91$ dyads finished the experiment, $52$ of which were included in the analysis ($104$  participants in total). 
	Data from $39$ dyads was discarded because of low communicative accuracy.

    \subsubsection{Procedure}
        
\begin{figure}[htb]
			\noindent
			\centering
			\includegraphics[width=0.95\columnwidth]{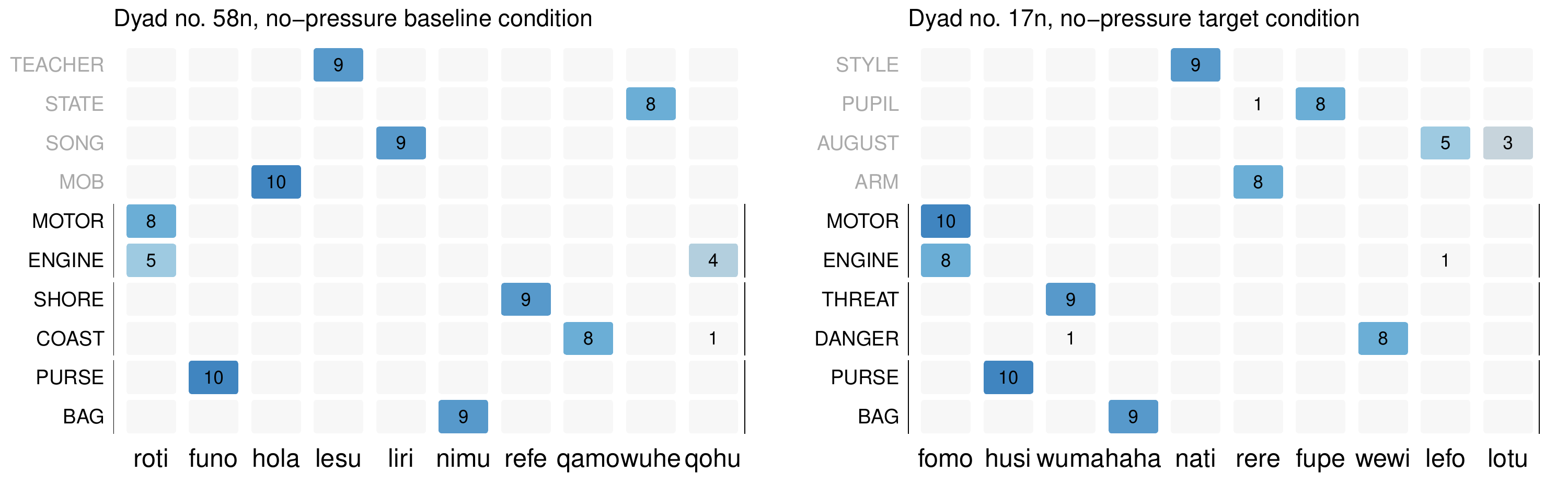}
			\caption{
				Signal-meaning matrices illustrating the no-pressure baseline (left) and target condition (right) of Experiment 4. Above average accuracy dyads are used again for visualization purposes, but their behavior is reflective of the average tendencies in this experiment.
				Here, participants make use of more signals in both conditions and generally colexify less than in previous experiments.
			}\label{fig_expm_matrices_extras}
	\end{figure}

    This setup is the same as that of Experiment 1 and 2: there is a baseline and a target condition, and the meaning space and its associated occurrence distributions are arranged as discussed in Section \ref{sec_expm_procedure}). The single difference is that participants are not forced to colexify any meanings, as the size of the signal space is 10, equal to the meaning space (illustrated in Figure \ref{fig_expm_matrices_extras}). 
         
    \subsection{Results}
    
    The results of applying the mixed effects logistic regression model to the data from Experiment 4 ($1306$ cases after operationalizing colexification) show that when the requirement to colexify is removed, by allowing a larger signal space, the difference between the baseline and target condition disappears ($p=0.37$, see Table \ref{table_regression4}; see also Figure \ref{fig_expm_all} back in Section \ref{sec_expm1} for a visualization across all conditions). 
    Communicative accuracy is roughly the same as in the other Mechanical Turk samples (recall Figure \ref{fig_accs}).
    
    \begin{table}[htb]
		\centering
		\begin{tabular}{rrrrr}
			\hline
		colexification \textasciitilde & Estimate & SE & $z$ & $p$ \\ 
			\hline
			intercept (no-pressure baseline condition, mid-game) & -0.73 & 0.27 & -2.74 & $<$0.01 \\
			+ condition (target) &                          0.24 & 0.36 & 0.68 & 0.5 \\
			+ round &                                       0.19 & 0.24 & 0.78 & 0.44 \\ 
			+ condition (target) $\times$ round &           -0.3 & 0.33 & -0.89 & 0.37 \\ 
			\hline
		\end{tabular}
		\caption{
			The fixed effects from the mixed effects regression model applied to data from Experiment 4, predicting the value of the colexification variable (reference level: ``no") by the interaction between condition (reference: baseline) and round number. Here, in contrast to previous experiments, the conditions yield similar results.
		}\label{table_regression4}
	\end{table}

		\subsection{Discussion}

\begin{figure}[htb]
			\noindent
			\includegraphics[width=1\columnwidth]{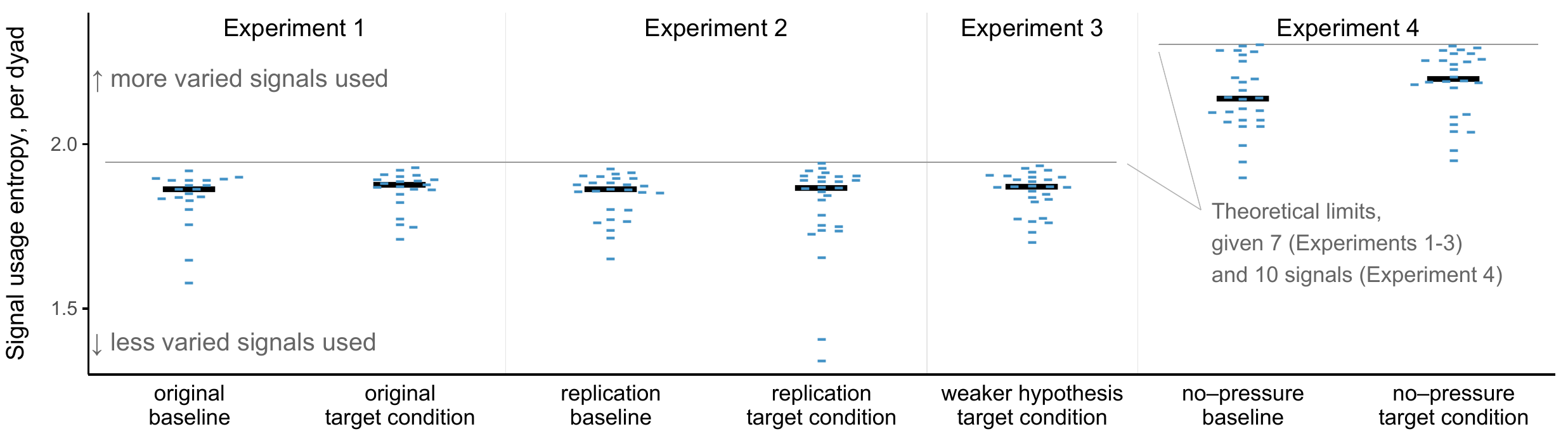}
			\caption{
				Signaling entropy across all dyads (blue notches), in the post-burn-in part of the game. Low values on the vertical axis indicate fewer signals were consistently used, higher values indicate a larger variety of signals were used by a given dyad. Overlapping values are pushed slightly aside horizontally to ensure visibility. The black bars represent medians. The gray vertical lines at 1.9 and 2.3 indicate maximal entropy given the size of the signal spaces.
				Dyads in Experiment 4 generally made use of most of the extended signal space (10 instead of 7), setting it apart from the rest of the experiments. 
			}\label{fig_entropy}
		\end{figure}

		When the signal space is larger, the difference between the baseline and the target condition disappears. 
		Participants behave in the baseline condition similarly to participants in the target conditions, avoiding colexification of the target pairs. 
		We also quantified signal usage entropy for all dyads across all conditions. The results show that in the no-pressure conditions with the expanded signal space,  in absolute terms more signals were used on average (Figure \ref{fig_entropy}). 
        In relative terms, signal usage in Experiment 4 was not that different from the previous experiments, with some dyads making use of most of the signal space (notches close to the limit lines in Figure \ref{fig_entropy}) and others being more conservative and colexifying instead.
		
		It appears participants favored the effort of remembering a few extra signals over having a simpler but more ambiguous language.
		While the signaling options are visible in the game at all times, the entire meaning space is never revealed all at once (recall Table \ref{table_expm_screens}). Participants here seem to have picked up on the relative abundance of the signals nevertheless, and unlike in the weaker hypothesis condition, participants did not colexify similar meanings to the same extent. 
		These results solidify the original findings of Experiment 1 and 2: the size of the signal space clearly makes a difference in participant behavior and allows for making inferences into speakers' lexification preferences. Naturally, our signal spaces are tiny compared to real lexicons that humans are able to memorize, not least because our signal space is designed for a lexicon that is to be conventionalized and used in the span of about half an hour. And like in natural languages, the complexity of the lexicon, the number of signals, has a limit. It is just that this limit is cognitive in nature in the real world and appears to be largely artificial (i.e. driven by our constraint on the size of the signal space) in our experiments.

		\section{General Discussion}\label{section_colex_discussion}
	
		Our research provides evidence that speakers' communicative needs affect their lexification choices, validating the viability of the mechanism hypothesized based on large-scale cross-linguistic studies \parencite[cf.][]{kemp_semantic_2018,xu_conceptual_2020}. Our research makes this connection explicit, and describes an experimental paradigm to test such hypotheses on the level of individual discourse --- in comparison to previous research focusing on the level of population consensus based on data such as dictionaries, grammars, and corpora \parencite[e.g.][]{ramiro_algorithms_2018,xu_conceptual_2020,mollica_grammatical_2020}.
		Below, we sketch some extensions to the experimental paradigm established here that we feel would be worthwhile to look into in order to gain a better understanding of the role of communicative need, similarity and associativity, and the formation of lexicons in general.

        \subsection{Extensions and implications}
        
		Future research could look into a number of aspects and parameters of the communicative need game, beyond the exploration in our two follow-up studies (Experiments 3 and 4).
		In terms of setup and procedure, we chose what we assumed would be a reasonably-sized meaning space for an experiment of this length, but this, as well as the length itself, are of course arbitrary, as are the chosen co-occurrence distributions that emulate the pressure of communicative need. It would be interesting to see both the effect of stronger and weaker pressures than employed here. 
		Need could also be gradually increased over the course of a longer game, to probe the strength of the pressure required for participants to modify an already established artificial communication system \parencite[which would loosely correspond to natural language users changing their language over their lifespan; cf.][]{sankoff_language_2018}.
		
		In this study, we chose direct similarity or synonymy as the semantic relationship to explore. \textcite[][]{xu_conceptual_2020} show that conceptual associativity (i.e. the \textit{car, engine} type) also correlates with cross-linguistic colexification patterns, roughly to the same extent. It would be interesting to use our paradigm to investigate whether dyads preferentially colexify based on associativity or similarity.
		Another possible predictor, related to associativity, could be co-occurrence probability (or mutual information), which could be inferred from a large text corpus and then tested experimentally.
		Differences between more fine-grained relationships like register-varying synonymy (\textit{abdomen, belly}) and hyponymy (\textit{bag, purse}) could also be probed, as well as how these preferences may correlate with historical patterns of sense formation and expansion \parencite[cf.][]{ramiro_algorithms_2018}.
		
		\textcite{xu_conceptual_2020} also discuss a potential role of frequency, namely that more commonly referred-to senses may be more likely colexified. We control for frequency in our experiment by making sure the occurrence distribution of meanings is uniform in all games. Future research could let frequency vary systematically to determine its importance.
		Previous research \parencite[cf.][]{atkinson_social_2018,raviv_larger_2019,segovia-martin_network_2020,raviv_what_2021,blythe_how_2021} has also demonstrated that community size and links between individuals has an effect on (artificial) language formation, learnability, and change. Another potentially interesting extension would be to run the colexification experiment using a larger speaker group than a dyad, perhaps also manipulating the connections between participants to investigate the proposed network effects.
		In our experiment, the participants were only presented with two possible meanings to guess between, to facilitate convergence on systematic associations in a short time frame. This also means it was trivial to achieve 50\% guessing accuracy. In a longer experiment, the number of choices could be increased.
		
		Our work may also have potentially interesting implications for historical linguistics. Population-level changes large enough to register on historical time scales must also start with differential utterance selection on the individual level, before they can compound over time and larger groups to become the norm \parencite[cf.][]{croft_explaining_2000,baxter_utterance_2006}. 
		If the mechanisms discussed and experimented with here are representative of utterance selection dynamics in natural languages, then the next interesting question would be: to what extent does communicative need constitute selection in language change? \parencite[cf.][]{andersen_structure_1990,baxter_utterance_2006,reali_words_2010, newberry_detecting_2017,steels_evolutionary_2018} . 
		A comprehensive understanding of lexical evolution, and language evolution in general, would benefit from merging the perspectives of individual mechanisms and population level consensus changes.

		\subsection{Complexity, information loss and communicative need}
		
		In more broad terms, our study interfaces with a growing body of work on the interplay between the orthogonal pressures of
		simplicity and informativeness in language evolution.
		The former relates to ease of learning; while
		the latter relates to low information loss or communicative cost
		\parencite[the terminology and foci vary between authors and disciplines, cf.][]{kirby_emergence_2002,gasser_origins_2004,kemp_kinship_2012,fedzechkina_language_2012,kirby_compression_2015,winters_languages_2015,carstensen_language_2015,beckner_emergence_2017,bentz_entropy_2017,nolle_emergence_2018,zaslavsky_semantic_2019,carr_simplicity_2020,smith_how_2020,steinert-threlkeld_ease_2020,haspelmath_explaining_2021,uegaki_nand_inprep,denic_complexityinformativeness_2021}. 
		These studies have yielded converging evidence that languages which are learned and used in communication --- the real-world ones, the artificial ones grown in the lab, as well as those evolved by computational agents --- all aspire to balance these two pressures, ending up somewhere along the optimal frontier. 
		
		Our results provide support for the argument that culture-specific communicative needs may modulate the location of a language on that frontier \parencite[cf.][]{kemp_semantic_2018}. The pressure for simplicity in lexicons can be relaxed in favor of more expressivity, given high enough communicative need (which we emulated in the target condition, and the no-pressure conditions) --- while informativeness can give way to simplicity when a less expressive lexical subspace does the job (cf. our baseline condition).
		
\begin{figure}[htb]
			\noindent
			\centering
			\includegraphics[width=0.5\columnwidth]{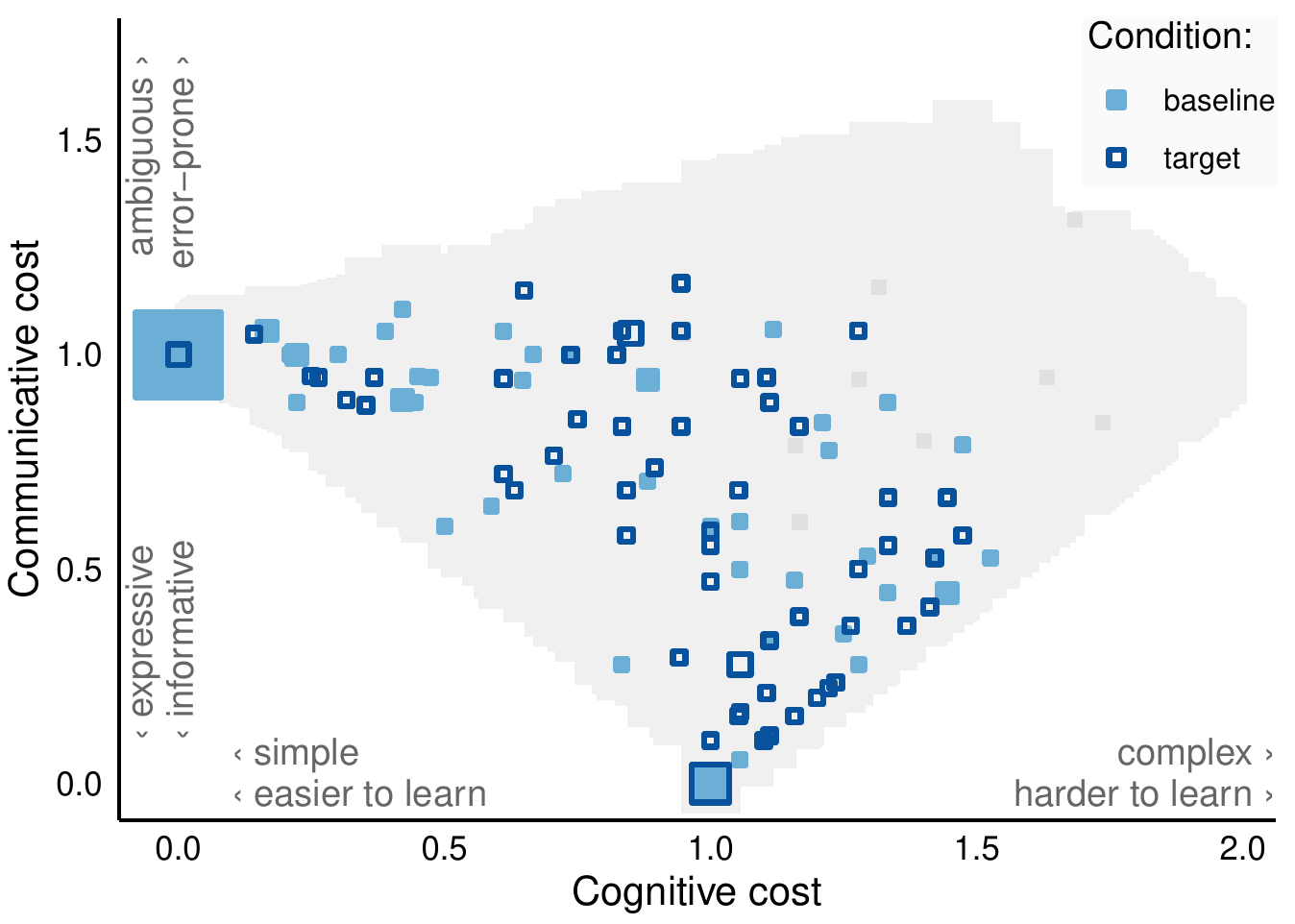}
			\caption{
				Average communicative cost and cognitive cost (complexity) scores in Experiment 1. Light blue squares are target meaning pairs (such as \textsc{drizzle-rain}) used by baseline condition dyads, dark blue ones are pairs as used by target condition dyads. Larger square size indicates multiple overlapping squares at those coordinates. Simulated results are depicted as gray blocks in the background. The few slightly darker gray squares, mostly in the top right, are the few dyads in Experiment 1 that performed below the minimal threshold of $59\%$.
				Most dyads communicate at or near the optimal points at $(0,1)$ and $(1,0)$, and none of the dyads (above the accuracy threshold) end up in the suboptimal top right corner. This picture mirrors findings in natural language lexicons, which also tend towards the optimal frontier (cf. Figure \ref{fig_infcompexample}).
			}\label{fig_expmresults_infcomp}
		\end{figure}

		To allow for systematic statistical testing of the hypothesis in the previous sections, we applied a transformation to the data (outlined in Section \ref{sec_exp_operationalize}) that operationalized colexification the most objective manner we could think of. Figure \ref{fig_expmresults_infcomp} is a reanalysis of data from Experiment 1 along the axes of complexity and expressivity, using an alternate coding scheme. To simplify things, we ignore the sender here and treat the language of each dyad as a collaborative effort. Each message containing a target meaning is assigned a simplified cognitive cost score and communicative cost score between $[0,2]$, which depends on the last most recent reference to the same meaning and the last reference to the synonym (target pair member) of the current meaning. Figure \ref{fig_expmresults_infcomp} displays the mean scores for each target meaning pair used by each dyad in Experiment 1, across all their respective utterances. 
		We also simulate the full possibility space of the results under this coding scheme, using a simple agent-based model (shown as gray blocks in Figure \ref{fig_expmresults_infcomp}). This is to provide meaningful dimensionality to graph: given the number of signals and meanings, it would be impossible to obtain maximal score simultaneously on both axes. The technical details of both procedures are further described in the Appendix.
		
		In summary, dyads tend towards the optimal frontier between complexity and expressivity (or in this case, the two optimal points at $(1,0)$ and $(0,1)$; positioning near the bottom left diagonal just indicates mixed strategies). Baseline dyads favor simplicity (having the luxury to do so), while target condition dyads sacrifice simplicity to reduce communicative cost (in response to our manipulation driving them to do so).

		\subsection{Beyond small subsystems}
		
		Extrapolating the communicative need argument beyond the grammatical and lexical subsystems mentioned in Section \ref{sec_intro} to the scale of entire languages, we would expect semantic spaces of different languages to be mostly uniform in density --- how many words are used to express shades of any given concept or meaning subspace --- but differ in exactly where culture-specific communicative needs of the time either require more detail, or where fewer words will suffice \parencite[analogous to uniform information density on the level of utterances; cf.][]{levy_communicative_2018}.	
		Previous research has focused on delimited domains of language like tense or kinship. This makes sense both from a data collection and computational point of view: quality cross-linguistic data is not trivial to acquire, and neither is computing complexity, information loss or expressivity, the larger the system under scrutiny. 
		This is then the next challenge: understanding these pressures and the evolution of lexicons and grammars, over time and cross-linguistically, on the scale of entire languages (as opposed to isolated domains). 
		This would require combining 
		explicitly quantified metrics of simplicity and expressivity \parencite[cf.][]{piantadosi_word_2011,bentz_entropy_2017,zaslavsky_semantic_2019,steinert-threlkeld_ease_2020,mollica_grammatical_2020},
		some estimate of communicative need \parencite[cf.][]{regier_word_2015,karjus_quantifying_2020}, 
		some measure of density or colexification \parencite[see Chapter 5.4 of][for one potential approach]{karjus_competition_2020},
		and if using a machine learning driven approach such as word embeddings, either a joint semantic model of multiple languages allowing for direct cross-linguistic comparison of lexical densities and colexification \parencite[e.g.][]{chen_unsupervised_2018,thompson_quantifying_2018,rabinovich_typology_2020},
		or language-specific diachronic semantic models to observe changes in colexification over time \parencite[e.g.][]{rosenfeld_deep_2018,dubossarsky_timeout_2019,ryskina_where_2020}.

		\FloatBarrier
		\section{Conclusions}
		
		We investigated the cross-linguistic tendency of colexification of similar concepts from earlier lexico-typological research 
		using artificial language experiments, and tested the hypothesis that colexification dynamics may be driven not only by concept similarity but also the communicative needs of linguistic communities. Our data supports both claims: speakers readily colexify similar concepts, unless distinguishing them is necessary for successful communication, in which case they do not. These results, despite being based on artificial communication scenarios and small lexicons, illustrate the interaction between similarity and communicative need in shaping colexification. We also proposed pathways for future study of these phenomena beyond small word sets and on the scale of entire lexicons.
		
		Language change is driven by a multitude of interacting forces, ranging from random drift to sociolinguistic pressures to institutional language planning, to selection by speakers for more efficient and expressive forms. 
		Our work supports the argument that speakers' communicative needs --- a factor balancing and modulating the relative importance of the higher-level pressures for simplicity and informativeness --- should be considered as one of such forces.

		\section*{Acknowledgements}
		
		We would like to thank Yang Xu, Barbara C. Malt and Mahesh Srinivasan for useful discussions and comments, Jonas Nölle for advice with the initial experimental design, and the anonymous reviewers of Cognitive Science as well as the associate editor, Padraic Monaghan, for their constructive feedback.
		
		\section*{Author contributions}
	
		Andres Karjus designed and carried out the experiments, conducted the analysis, wrote the text, and created the figures.
		Tianyu Wang carried out additional experiments.
		Kenny Smith, Richard A. Blythe and Simon Kirby provided advice on the design of the experiment and data analysis, as well as edits and comments on the text.
		A shorter version of this paper formed a part of a chapter in the doctoral thesis of the first author \parencite{karjus_competition_2020}. 
		
		\section*{Funding}
		Data collection for this research was funded by the Postgraduate Research Support Grant of the School of Philosophy, Psychology and Language Sciences of the University of Edinburgh.
		This research also received funding from the European Research Council (ERC) under the European Union’s Horizon 2020 research and innovation program (Grant Agreement 681942), held by Kenny Smith.
		Andres Karjus is supported within the CUDAN ERA Chair project for Cultural Data Analytics at Tallinn University, funded through the European Union Horizon 2020 research and innovation program (Project No. 810961), and was also supported by the Kristjan Jaak postgraduate scholarship of the Archimedes Foundation of Estonia during initial data collection.

		\section*{Code and data availability}
		
		All the experiment data and scripts to replicate the analyses are available in the following Github repository, 
		along with the full codebase of the Shiny game application we developed to run the dyadic experiments: \\ \mbox{\urlstyle{rm}\url{https://github.com/andreskarjus/colexification_experiment}}

		\FloatBarrier

\begingroup
\setlength{\emergencystretch}{8em}
\printbibliography
\endgroup

\FloatBarrier

\section*{Appendix: Details on the complexity-ambiguity calculation}

	The procedure to produce Figure \ref{fig_expmresults_infcomp} in Section \ref{section_colex_discussion} is the following. Each message produced by a dyad after the burn-in period, which contains a target meaning, is assigned a cognitive cost (complexity) score and communicative cost (ambiguity) score. As a simplification, we consider the results by dyads, ignoring who sent a given message within a dyad. Only messages containing target meanings are scored, as distractor meanings lack synonyms in the meaning spaces (of Experiment 1, which we analyze here).
	
	The cognitive cost score is set to 0, if a given utterance does not increase the complexity of the language, within the target pair: if the last reference to the same meaning used the same signal, and the last reference to the synonym (target pair member) of the current meaning also used the same signal. Using a different signal or distinguishing the current meaning from its synonym increases complexity by 1 point each. 
	
	Communicative cost is scored as 0 if the last reference to the same meaning used the same signal, and the last reference to the synonym of the current meaning used a different signal. Using a different signal or colexifying the current meaning both increase ambiguity (communicative cost, chance of misinterpreting), so doing either costs 1 point each. Given 7 signals and 10 meanings, some meanings are bound to be colexified. The minimal sum of these two scores for any given utterance is 1: it is impossible to be simultaneously maximally simple and maximally informative. The highest sum of scores is 3, which may result in random assignments of signals to meanings, or intentionally misleading behavior (see Table \ref{table_appendix_scores} for an example).
	
    \begin{table}[htb]
		\centering
		\begin{tabular}{rrrrrr}
	        current meaning & last \textsc{task} & last \textsc{job} & new signal &  complexity & ambiguity \\ 
			\hline
	        \textsc{task} & nopo & nopo & nopo & 0 & 1 \\
	        \textsc{task} & nopo & mumi & nopo & 1 & 0 \\
	        \textsc{task} & nopo & nopo & mumi & 1 & 1 \\
	        \textsc{task} & nopo & mumi & mumi & 1 & 2 \\
	        \textsc{task} & nopo & mumi & fita & 2 & 1 \\
			\hline
		\end{tabular}
		\caption{
			An example of scoring a message in an experiment on the axes of complexity (cognitive cost) and ambiguity (communicative cost). Given a round where the meaning is \textsc{task}, the scores depend on the last most recent lexification of both \textsc{task}, its target pair member, \textsc{job}, and the new signal used to express \textsc{task}. Using different signals to distinguish the two meanings increases complexity but decreases ambiguity. The sum of these scores is the highest if the participants use entirely different signals for the meanings (last row) or if they change the signals without converging on stable associations (second to last row).
		}\label{table_appendix_scores}
	\end{table}
	
	In addition to re-analyzing the results of our human experiments, we implement a simple agent-based model that replicates our experimental setup (up to 7 signals, equally frequently occurring 10 meanings; 135 rounds). We use the results from this model --- analyzed using the same coding procedure --- to provide meaningful dimensionality to the human results in Figure \ref{fig_expmresults_infcomp}.
	The agent-based model is constructed as a single-agent system, representing a dyad, which plays a simple naming game. The model has two parameters: the number of signals allowed (1--7) and the naming strategy. At each round, the agent assigns a signal to a given meaning. In the most simple case, it only ever produces a single signal, leading to a ``degenerate" language \parencite[see][]{kirby_compression_2015}. Other strategies include:
	\begin{itemize}[topsep=0pt]
	\setlength\itemsep{0em}
      \item random signaling;
      \item fixed assignment to as many meanings as possible, with perfect memory (and random assignment for meanings for which there are not enough signals);
      \item fixed assignment with perfect memory, but colexifies pairs of meanings (others assigned randomly);
      \item fixed assignment with perfect memory, but colexifies pairs of meanings (others assigned randomly, but avoiding colexification with the pairs where enough signals available);
      \item rational \parencite[in the sense of][]{frank_predicting_2012}, keeping the entire lexification history in memory;
      \item rational, but keeping only the most recent lexification;
      \item intentionally misleading, i.e. the inverse of the rational strategies;
      \item and also combinations of the above.
    \end{itemize}
    Each combination of these parameters is repeated 20 times.
	The results are analyzed exactly the same way as the human experiments (including the designation of a ``burn-in" period), and are graphed as gray background blocks in Figure \ref{fig_expmresults_infcomp}.

\end{document}